# Dispelling Classes Gradually to Improve Quality of Feature Reduction Approaches

Shervan Fekri Ershad and Sattar Hashemi

Department of computer science, engineering and IT, Shiraz University
Shiraz/Iran
*shfekri@shirazu.ac.ir*
*S_hashemi@shirazu.ac.ir*

## ABSTRACT

*Feature reduction is an important concept which is used for reducing dimensions to decrease the computation complexity and time of classification. Since now many approaches have been proposed for solving this problem, but almost all of them just presented a fix output for each input dataset that some of them aren't satisfied cases for classification. In this we proposed an approach as processing input dataset to increase accuracy rate of each feature extraction methods. First of all, a new concept called dispelling classes gradually (DCG) is proposed to increase separability of classes based on their labels. Next, this method is used to process input dataset of the feature reduction approaches to decrease the misclassification error rate of their outputs more than when output is achieved without any processing. In addition our method has a good quality to collate with noise based on adapting dataset with feature reduction approaches. In the result part, two conditions (With process and without that) are compared to support our idea by using some of UCI datasets.*

## 1. INTRODUCTION

Since the middle of 20[th] century when artificial intelligence science was established, classification methods were too important. By the pass of time, datasets that were used for classification got more complex than past such as geographic or discovery dataset. One of the items that increase the complexity of dataset is the number of dimensions (features). So a new concept called feature reduction was inducted in the literature of AI (Fukunnaga 1991) [4].

Fukunnaga said, If we describe input data as a matrix like X= $\{x_1 \ldots x_n\} R^{m*n}$, where "n" is the sample number and "m" is the original feature dimensions, So the purpose of linear feature extraction is to search for a projection matrix like W$\in R^{m'*m}$ that transforms $x_i \in R^m$ into a desired low-dimensional representation $y_i \in R^{m'}$, where $m' \ll m$ and $y_i = W x_i$.

Typically, the projection matrix W is learnt by optimizing a criterion describing certain desired or undesired statistical or geometric properties of the data set. Different criterions lead to different kinds of linear feature exaction algorithms. Among them, Principal Component Analysis (PCA) (Joliffe, 1986) and Linear Discriminant Analysis (LDA) (Fukunnaga, 1991) have been the two most popular ones owing to their simplicity and effectiveness. Another popular technique called Locality Preserving Projections (LPP) (He & Niyogi, 2004) [5], has been proposed for linear feature extraction by preserving the local relationships within the data set. In (Yan et al., 2007) [1], many classical linear feature extraction techniques are unified into a common framework known as Graph Embedding. To avoid the high time and memory usage associated with eigenvalue decomposition in LDA, the Spectral Regression Discriminant Analysis (SRDA) (Cai et al., 2008) [2], was proposed based on ridge regression.

Now the output dataset Y is in a low-dimensional so the complexity of classification Y and the time of that are less. Also, computation complexity of Y for any other cases is less than X. It's easy to know that feature extraction wastes some pieces of the dataset's information. All of the methods for feature extraction are common in this object, so algorithms try to find a better formula for matrix W to decrease the amount of waste information to decrease the miss classification error rate of output(Y). Almost none of the previous methods have a good generalization for each kind of different datasets. To solve this problem, in this paper we present an approach to adapt the input dataset with every feature extraction methods. First we present a new concept called Dispelling Classes Gradually (DCG) to increase severability and separability of classes based on their labels and then we process the original input dataset with DCG and create a new dataset as an input dataset of the feature extraction methods to decrease the amount of misclassification error rate on its output rather than original situation.

### 1.1. PAPER ORGANIZION

The reminder of this paper is organized as follows: Section two is related to the description of Dispelling Classes Gradually technique (DCG) and the way of this estimation. In the section three the proposed approach is described. Section four has justification about quality of proposed approach. In section five necessary parameters are tuned. Section six is related to the description of proposed approach's noise insensitivity. Finally, the results and conclusion included.

## 2. DCG CONCEPT

In the numerical datasets, misclassification error rate decreases with increasing the distance between classes[14]. So it's necessary and beneficial to increase these distances. In the basic mathematics it's determined that if two different numbers like $P_1$ and $P_2$ be subtracted two another fix numbers like $Q_1$ and $Q_2$, respectively, and this action get repeated for α times, where $Q_1 < Q_2$, it's authenticate that in extreme, the distance between $P_1$ and $P_2$ will be more than first generation. This principal is shown in figure (1).

If we assume class labels to numeric labels and then assign $Q_i$ for class labels, separability of classes increases at extreme situation by subtracting each class label from its sample value in available features. This subject is described in (1).

$$(X_i - \alpha L_{X_i}) - (X_j - \alpha L_{X_j}) > X_i - X_j \qquad (1)$$

Where, $X_i$ is the $i_{th}$ sample and $X_j$ is the $j_{th}$ sample and $L_{X_i}$ is the class label of $X_i$ and $L_{X_j}$ is the class label of $X_j$. Figure (2) shows the results of applying (1) on Iris (UCI Dataset) on two features.
There is a same authentication for negative numbers of α. If the numbers of α be negative just changes the orientation of classes move, but result of applying DCG is same as positive α values. According to figure (2) and formula (1), the distances between all samples of each class doesn't change also variance of each class, but separability of classes increase by dispelling mean of the classes.

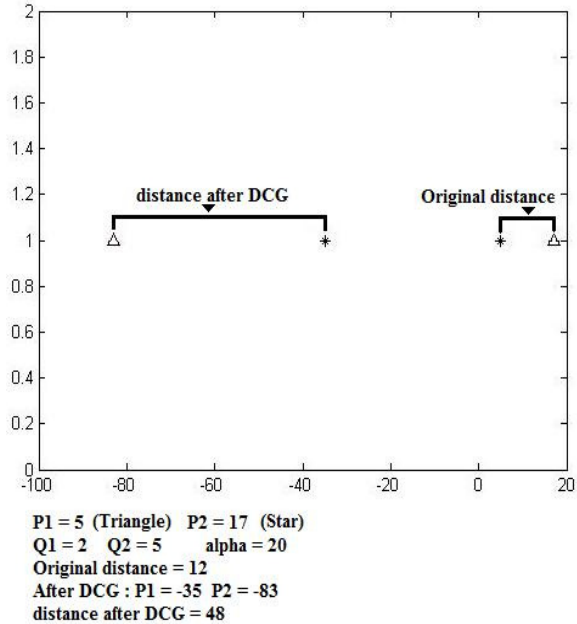

P1 = 5 (Triangle)   P2 = 17  (Star)
Q1 = 2   Q2 = 5       alpha = 20
Original distance = 12
After DCG : P1 = -35   P2 = -83
distance after DCG = 48

**Figure 1.** Description of DCG

## 3. PROPOSED APPROACH

According to previous part, DCG is too beneficial to increase separability of classes. But DCG does its process based on class labels. So on test set, DCG is not useful because we don't know sample labels. But for train set, its may be useful. According to introduction part, all of feature extraction methods try to describe an accurate approach to create projection matrix W based on train set. So if DCG increases separability of train set before using to create projection matrix and then dimension reduced output dataset is compute based on projection matrix, then misclassification error rate of output decreases and output set Y classifiers easily get more than output of original input set.

To implement this method, sample labels of train set assume to be represented as a matrix C= $[c_1 \ldots c_n] \in R^{N*N_c}$ , where N is the number of samples and $N_c$ is the number of labels and the elements of the indicator vector $C_i$ is set to be 1 to $N_c$ . Then do $(X - C)$ for $\alpha$ times. This method has 3 important advantages. (A) Adapting dataset with feature extraction method (B) the results never are worse than original manner. (C) Increasing the noise tolerability of dataset.

(A) According to introduction part, if the input dataset of feature reduction be original just there is a fix output. The misclassification error rate of this fix output probably won't be desired. But with using our method it's possible to adapt input set with feature reduction method more than primitive manner. By using optimal $\alpha$ it's possible to create a better output with lower misclassification error rate than original output.
(B) If the number of α be zero, original input and processed input are same so the feature extraction's outputs of them are same and there is not any difference between misclassification error rates.
(C) One of the motivations that is presented for feature extraction methods is noise tolerability. In almost the entire feature extraction methods if noise sits on input dataset, the misclassification error rate of output gets more than original manner. But our method can cover this motivation to a large extent. Our method does this action based on two features that are hidden in its formula.

First feature is α. If the input dataset is noisy we can compute the optimal α to adapt feature reduction method with noisy dataset. It's useful to calculate an output with error rate close the situation that input set is not noisy. Second feature is C. one of the methods that is presented to decrease the effect of noise is using fuzzy. In matrix C there is a concept like fuzzy which is hidden. When we subtract sample's values of their labels in reality we subtract sample's values of all of the labels just with different coefficient. Because all of labels are coefficients for the other ones.

$$X_i = X_i - \alpha L_{X_i} \quad , \quad X_j = X_j - \alpha L_{X_j}$$

$$\text{If} \quad L_{X_i} = \beta L_{X_j} \quad \rightarrow \quad X_i = X_i - \beta \alpha L_{X_j} \qquad (2)$$

To understand better, the flowchart of proposed approach is shown in figure3. Also it is compared with popular feature reduction approaches without train set processing.

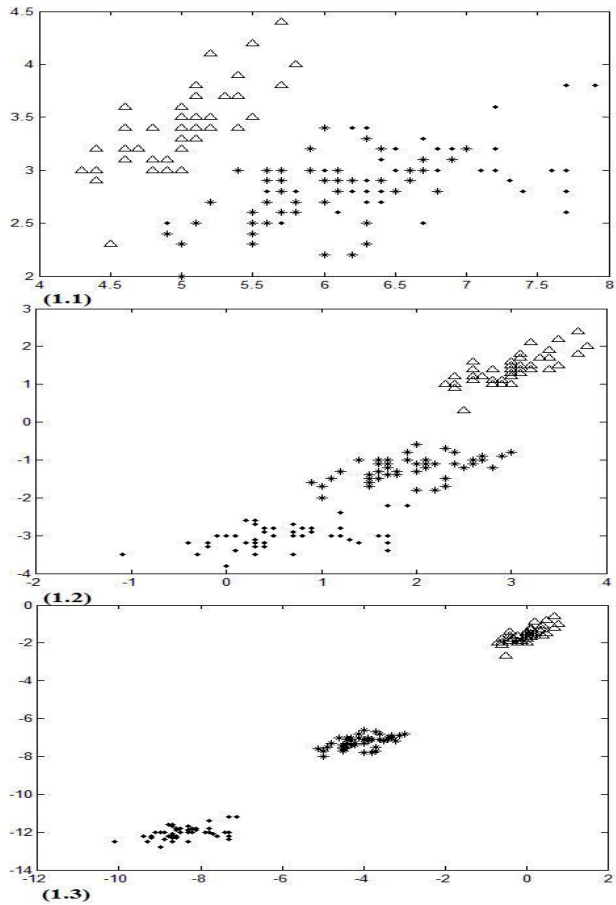

**Figure. 2.** Effect of DCG on first and second features of Iris dataset
(1.1) Original (1.2) α=2 (1.3) α=5

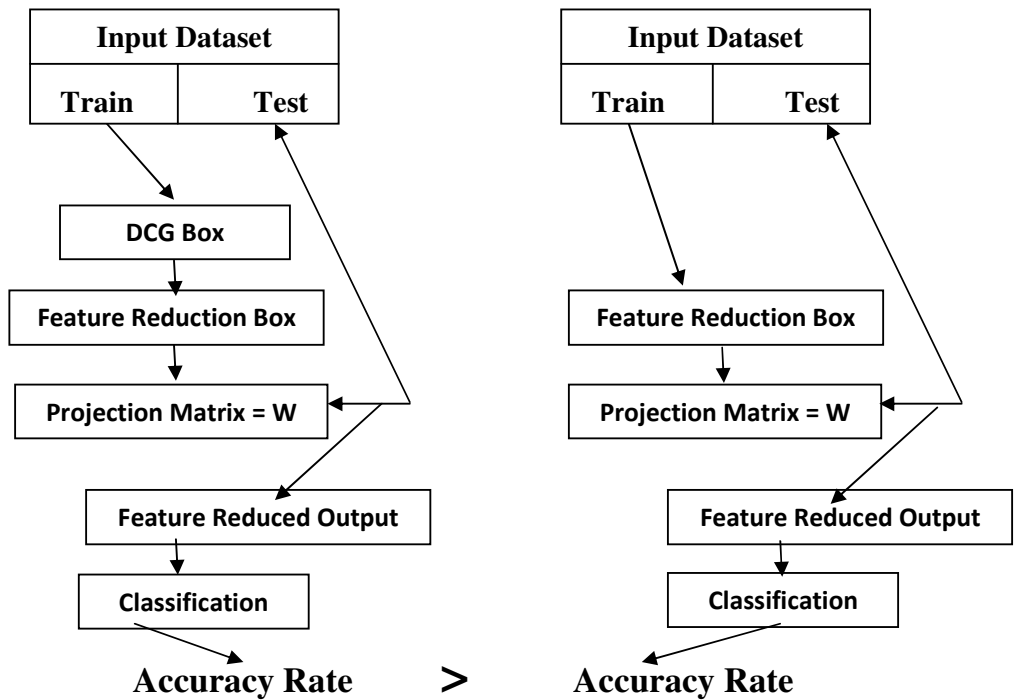

**Figure 3.** Flowchart of Comparing using DCG and Classical feature reduction

## 4. JUSTFICATION

The main question which is mentioned in this article is that why applying DCG concept on input dataset could increase the quality of feature reduction approaches. This question will be answered true this section. Almost all feature reduction approaches purpose is to describe new dimensions which have two below mentioned signification based on finding the mean, variance, and eigenvectors of each class in input dataset.

(A) The first signification is that new dimensions are less than original dimensions
(B) The second signification is that separability of classes ability is saved as far as possible.

As described in section two after processing the input dataset by DCG, both slope of each eigenvector and variance of each class remain constant and this happened because the transfer amount of each sample is equal to transfer amount of the same class's samples. Consequently, the main section of feature reduction approaches which is finding slope of eigenvectors, and variance of classes will be remained constant and only mean of each class will be changed. This subject is shown in figure (4). As it is shown in figure (4), the slope of eigenvectors of dataset is not changed after processing the dataset by DCG.

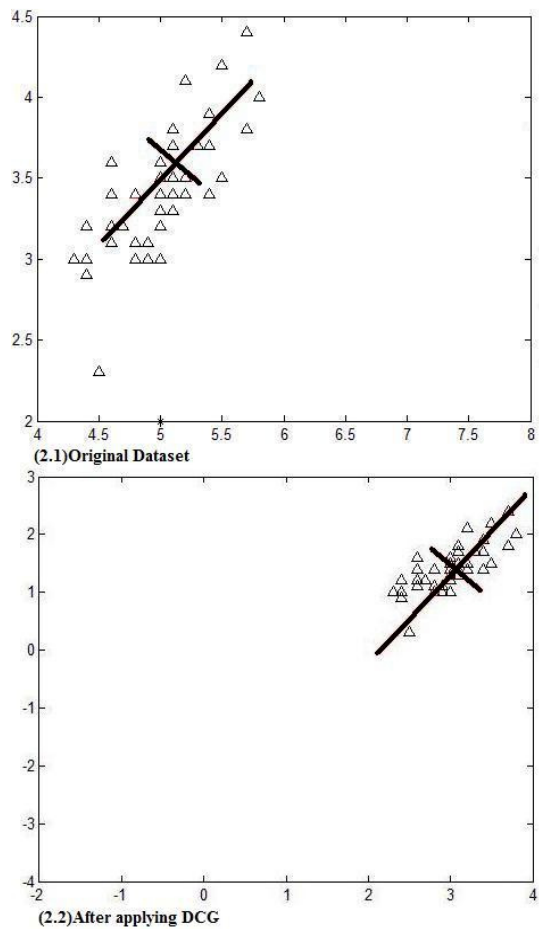

**Figure4.** Effect of DCG on Mean and eigenvectors of one of the classes of iris dataset

According to proposed approach, it is asserted by transferring coordination of different classes's mean, and distance between mean, it is possible to find the new coordination for each means and in this situation classification accuracy rate of feature reduction approach output is higher than original input dataset. Briefly, this article is mainly trying to show that by using DCG for processing input dataset, the feature reduction output error rate will be less than when input dataset is not processed.

## 5. PARAMETER SELECTION

To process input dataset of feature extraction methods there is just an input parameter. The number of DCG Loop called α is an important parameter. A should tune correctly to fulfill our aims. To select α there are two important points to pay attention. (I) according to DCG part, DCG avouch the separability of classes in infinite situation .So there are some values for α at each dataset that α doesn't dispel classes enough and it's possible that some of its values decrease the separability of classes. Because the move orientations of classes for dispelling are hidden, maybe the move orientation of a class is like another one and of course their speeds are never the same, so for some of α values, it's possible that classes come near each other and their distances are less than primitive situation. This range of α is called loop's problem maker range (LPMR), so we need the values that are out of this range. (II) But the second point to select α refers to feature extraction method. Using DCG method to process input datasets certainly changes the values of samples in each feature so it's important that these transforms don't cause

Advanced Computing: An International Journal ( ACIJ ), Vol.3, No.3, May 2012

problem for feature extraction algorithms. Because in some of the feature extraction algorithms there are some formula that don't work correctly with every values.

For example one of the new feature extraction methods that is presented in ICML 2009 by Xiao-Tong and Bao-Gang is REDA [8]. This method calculates the best projection matrix W in an iterative manner. In their algorithm there is a formula that computes an item by this form: exp (-$\|Wx - C_i\|^2/\sigma^2$). we know exponential of numbers of more than 30 is too huge and less than -13 is considered as zero for all cases. So it's important that new value of X after process don't be out of this bound. For solving this problem we can compute the good range of α by (3).where $m_i$ is the minimum value in samples of $i^{th}$ sample's label

$$\theta_{min} < (W(m_i - \alpha L_{X_i}) - C_i)^2/\sigma^2 < \theta_{max} \qquad (3)$$

Because of using just input dataset, there aren't any other parameters which are affected on accuracy rate of proposed approach.

According to these points and basic theory of the DCG concept, the misclassification error rates of feature extraction's outputs are different for different numbers of α. In the table (1) for example we computed the accuracy of the SRDA's outputs of the input dataset (Haber-man) after applying DCG based on numbers of α between numbers 1 to 30.

**Table 1.** Feature extraction output's accuracy after applying DCG based on different α

| Haber-Man(after Applying DCG) | | | |
|---|---|---|---|
| SRDA | | | |
| Number of loops | KNN (Accuracy ± std-dev) | Number of loops | KNN (Accuracy ± std-dev) |
| α = 1 | 64.27 | α = 16 | 63.89 |
| α = 2 | 64.06 | α = 17 | 64.29 |
| α = 3 | 65.09 | α = 18 | 64.80 |
| α = 4 | 64.32 | α = 19 | 65.09 |
| α = 5 | 64.68 | α = 20 | 65.87 |
| α = 6 | 64.27 | α = 21 | 66.37 |
| α = 7 | 64.05 | α = 22 | 66.10 |
| α = 8 | 63.98 | α = 23 | 66.13 |
| α = 9 | 63.77 | α = 24 | 66.30 |
| α = 10 | 63.87 | α = 25 | 65.43 |
| α = 11 | 63.69 | α = 26 | 65.80 |
| α = 12 | 63.99 | α = 27 | 65.43 |
| α = 13 | 63.76 | α = 28 | 64.80 |
| α = 14 | 64.23 | α = 29 | 65.20 |
| α = 15 | 63.67 | α = 30 | 65.06 |

According to the table (1), we plan the figure (5) based on numbers of α Between 1 to 60. According to the figure (5), the amount of the feature extraction's outputs accuracies has tolerance, so there are some local optimums and a global optimum. The authors suggest hill climbing algorithms or simple genetic algorithm (J.Holland 1970's) with a fitness function based on output set's misclassification error rate to compute best value of α (Global optimum).

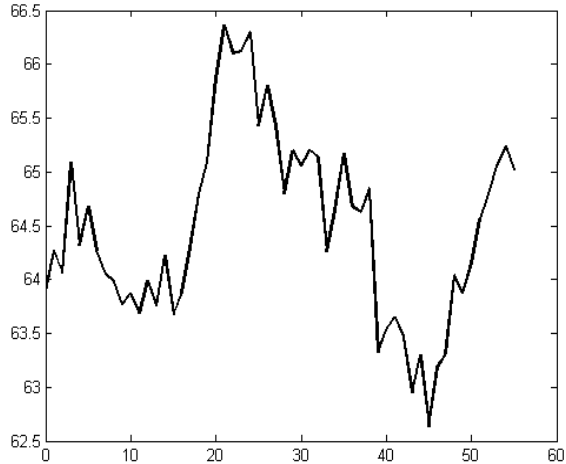

**Figure.5.** Tolerance of feature extraction output's accuracy based on different α

## 6. NOISE INSENSITIVITY

One of the most advantages of the proposed approach is noise insensitivity. According to different problems that occur, some of numeric values in dataset will change, it called noise. So, by occurring noise the quality of feature reduction approaches certainly decrease. By using the proposed approach, after doing some of DCG loops, the noise samples will nearing to another samples which have same labels, so, the effect of noise samples in the performance of feature reduction approaches will decrease.

## 7. RESULTS

In this part the misclassification error rate of our idea and original methods are compared. In each table of this part, first we computed the misclassification error rate of feature extraction's output of one of the original UCI datasets. It is shown in second columns of each row. The feature reduction approaches which are used are SRDA [2], PCA [1] and REDA-SRDA [8]. After that, the DCG is applied on the input datasets. So, the feature reduction approaches are applied on their processed input datasets and next classified. Misclassification rates are shown in fourth columns. Also, Parameter α is computed with a SGA (simple genetic algorithm) for each datasets and highlighted in third columns of each rows. All of the misclassification error rates computed with k-nearest neighborhood classifier [12] and we tried to use best number of k for each dataset to decrease the error rate. The results show, that the proposed approach increases the quality of feature reduction methods. Also, to prove the quality of proposed approach, the SVM classifier was used based on "one against all" technique, to compare, which is shown in the tables 6 & 7.

**Table 2.** Performance comparison on Huber-Man

| Huber-Man (UCI Dataset) |||||
|---|---|---|---|---|
| **Original Input Dataset** || **Input Dataset after Appling DCG** |||
| **Classifier / Methods** | **KNN : Accuracy ± std-dev** | **Classifier / Methods** || **KNN : Accuracy ± std-dev** |
| PCA | 70.45±0.17 | PCA | α=6 | 71.42±0.28 |
| SRDA | 63.89±0.24 | SRDA | α=22 | 66.20±0.30 |
| REDA-SRDA | 64.75±0.29 | REDA-SRDA | α=28 | 66.70±0.37 |

<pre>                     Advanced Computing: An International Journal ( ACIJ ), Vol.3, No.3, May 2012</pre>

Table 3. Performance comparison on Breast-Cancer

| Breast-Cancer (UCI Dataset) | | | |
|---|---|---|---|
| **Original Input Dataset** | | **Input Dataset after Appling DCG** | |
| Classifier / Methods | KNN : Accuracy ± std-dev | Classifier / Methods | KNN : Accuracy ± std-dev |
| PCA | 80.86±0.19 | PCA     α=5 | **87.06±0.15** |
| SRDA | 88.69±0.19 | SRDA    α=0 | **88.69±0.19** |
| REDA-SRDA | 91.34±0.50 | REDA-SRDA α=14 | **92.15±0.20** |

Table 4. Performance comparison on Glass

| Glass (UCI Dataset) | | | |
|---|---|---|---|
| **Original Input Dataset** | | **Input Dataset after Appling DCG** | |
| Classifier / Methods | KNN : Accuracy ± std-dev | Classifier / Methods | KNN : Accuracy ± std-dev |
| PCA | 64.46±0.32 | PCA     α=2 | **65.52±0.24** |
| SRDA | 56.69±0.25 | SRDA    α=57 | **63.23±0.20** |
| REDA-SRDA | 65.58±0.20 | REDA-SRDA α=0 | **65.58±0.20** |

Table 5. Performance comparison on Lung-Cancer

| Lung-Cancer (UCI Dataset) | | | |
|---|---|---|---|
| **Original Input Dataset** | | **Input Dataset after Appling DCG** | |
| Classifier / Methods | KNN : Accuracy ± std-dev | Classifier / Methods | KNN : Accuracy ± std-dev |
| PCA | 35.29±0.30 | PCA     α=2 | **42.22±1.40** |
| SRDA | 52.23±1.30 | SRDA    α=47 | **54.50±0.90** |
| REDA-SRDA | 49.36±0.60 | REDA-SRDA α=76 | **53.79±0.70** |

(1) Haber-Man (306 instances 3 attributes 2 classes )
(2) Breast-Cancer(699 instances 10 attribute 2 classes)
(3) Glass (214 instances 9 attribute 7 classes)
(4) Lung-Cancer(32 instances 56 attributes 3 classes)

<pre>                                                                                            103</pre>

**Table 6.** Performance comparison based on SVM on Haber-Man and Breast-Cancer

| Dataset<br>Methods | Haber-Man | | Breast-Cancer | |
| --- | --- | --- | --- | --- |
| | Original | After DCG | Original | After DCG |
| PCA | 73.22±0.23 | **76.65±0.53** | 81.32±0.37 | **89.05±0.41** |
| SRDA | 70.31±0.12 | **75.42±0.27** | 89.32±0.26 | **89.32±0.26** |
| REDA-SRDA | 68.63±0.45 | **71.92±0.18** | 92.05±0.41 | **94.11±0.72** |

**Table 7.** Performance comparison based on SVM on Glass and Lung-Cancer

| Dataset<br>Methods | Glass | | Lung-Cancer | |
| --- | --- | --- | --- | --- |
| | Original | After DCG | Original | After DCG |
| PCA | 66.52±0.14 | **67.02±0.27** | 43.82±0.76 | **51.74±0.64** |
| SRDA | 53.62±0.82 | **62.77±0.61** | 56.72±0.38 | **60.11±0.27** |
| REDA-SRDA | 65.58±0.20 | **66.43±0.59** | 52.04±0.60 | **56.03±0.54** |

## 8. CONCLUSION

In this paper we presented a new concept call DCG to increase separability of classes. DCG is provided by subtracting features values of their labels. So we processed the input dataset of feature extraction methods based on DCG. The results showed that output of feature reduction methods have misclassification error rate when the input is processed less than original dataset. Next we presented some ways to compute optimum numbers of DCG's loops and we mentioned some motivations for that. The results showed the quality of our idea based on various classifications. Some of advantages of proposed approach are:
1) Suitability with near all of the classifiers based on processing stage
2) Noise insensitivity at the result of using labels
3) Low computational complexity against some of previous approaches
4) The accuracy rate results never be lower by using DCG than when original dataset are used

## 9. FUTURE WORK

According to the fifth section, one interesting future research direction is to study how to compute numbers of DCG's loop without any algorithms just with formula. It will decrease the time and computation complexities.